%% file: main.tex
\documentclass[conference]{IEEEtran}
\usepackage{cite}
\usepackage{amsmath,amssymb,amsfonts}
\usepackage{algorithmic}
\usepackage{graphicx}
\usepackage{textcomp}
\usepackage{xcolor}

\usepackage{orcidlink}
\usepackage{booktabs}
\usepackage{makecell}
\usepackage{multirow}
\usepackage{caption}

\newtheorem{definition}{Definition}
\def\BibTeX{{\rm B\kern-.05em{\sc i\kern-.025em b}\kern-.08em
    T\kern-.1667em\lower.7ex\hbox{E}\kern-.125emX}}
\begin{document}

\title{Bikelution: Federated Gradient-Boosting for Scalable Shared Micro-Mobility Demand Forecasting}

\author{%
    \IEEEauthorblockN{Antonios Tziorvas \orcidlink{0009-0005-0037-6264}}
    \IEEEauthorblockA{%
        \textit{Department of Informatics} \\
        \textit{University of Piraeus}\\
        Piraeus, Greece \\
        atzio@unipi.gr
    }
    \and
    \IEEEauthorblockN{Andreas Tritsarolis \orcidlink{0009-0009-9433-109X}}
    \IEEEauthorblockA{%
        \textit{Department of Informatics} \\
        \textit{University of Piraeus}\\
        Piraeus, Greece \\
        andrewt@unipi.gr
    }
    \and
    \IEEEauthorblockN{Yannis Theodoridis \orcidlink{0000-0003-2589-7881}}
    \IEEEauthorblockA{%
        \textit{Department of Informatics} \\
        \textit{University of Piraeus}\\
        Piraeus, Greece \\
        ytheod@unipi.gr
    }
}

\maketitle

\begin{abstract}
    The rapid growth of dockless bike-sharing systems has generated massive spatio-temporal datasets useful for fleet allocation, congestion reduction, and sustainable mobility. Bike demand, however, depends on several external factors, making traditional time-series models insufficient. Centralized Machine Learning (CML) yields high-accuracy forecasts but raises privacy and bandwidth issues when data are distributed across edge devices. To overcome these limitations, we propose Bikelution, an efficient Federated Learning (FL) solution based on gradient-boosted trees that preserves privacy while delivering accurate mid-term demand forecasts up to six hours ahead. Experiments on three real-world BSS datasets show that Bikelution is comparable to its CML-based variant and outperforms the current state-of-the-art. The results highlight the feasibility of privacy-aware demand forecasting and outline the trade-offs between FL and CML approaches.
\end{abstract}


%% Add paper sections 
\input{1.Introduction.tex}
\input{2.RelatedWork.tex}
\input{3.ProposedApproach.tex}
\input{4.ExperimentalSetup.tex}
\input{5.Conclusions.tex}

\section*{Acknowledgment}
    This work was supported in part by the EU Horizon Framework Programme under Grant Agreement No. 101093051 (EMERALDS; \url{https://www.emeralds-horizon.eu/}) and EU Horizon Europe R\&I Programme under Grant Agreement No. 101070416 (Green.Dat.AI; \url{https://greendatai.eu}).

%% Define the bibliography file to be used
\bibliography{99.References.bib}

\end{document}

%% file: 1.Introduction.tex
\section{Introduction}
    The rapid expansion of dockless bicycle-sharing systems (BSS) in urban environments has generated massive spatio-temporal data. These data enables municipalities and operators to optimize fleet allocation, alleviate traffic congestion, and promote sustainable mobility, thereby turning urban transportation into a highly dynamic, data-driven process \cite{mohamed2023}. 
    
    Bike Demand Forecasting (BDF) is a critical task for operational planning in the urban domain. Accurate BDF is a non-trivial task since demand is affected by spatial relationships, such as rider flows between neighboring stations, the topology of the road network, and the geometry of the underlying dock-less service area \cite{DBLP:journals/tits/LiL24a}. Conventional time-series models and shallow-learning predictors quickly become inadequate because they cannot capture these intertwined spatio-temporal dependencies.
    
    Recent advances in Graph Neural Networks (GNN) have shown that explicit modeling of the BSS topology yields state-of-the-art forecasting performance on benchmark datasets \cite{XuSTMGCN2019,DBLP:journals/ijgi/LuoSYGFY22}. However, despite their predictive power, most GNN-based approaches are trained in a centralized manner, since all participating parties must pool raw trip records and corresponding BSS network topology to a single entity. This requirement directly conflicts with privacy regulations such as the EU General Data Protection Regulation (GDPR) and the California Consumer Privacy Act (CCPA), and it also threatens commercial confidentiality when multiple providers own disjoint sub-graphs of the overall BSS network.
    
    Horizontal Federated Learning (HFL) offers a principled solution to these constraints. Each participant trains a local copy of the model in-situ and exchanges only model updates with a centralized aggregation server. This paradigm preserves privacy and data sovereignty while while enabling a consistent forecasting model that benefits from the collective knowledge of all participants \cite{DBLP:conf/aistats/McMahanMRHA17,DBLP:journals/tsusc/PavlidisPYWMEKMD25}.
    
    Adapting GNNs to a federated setting, however, is technically challenging. Graph parameters are tightly coupled with topology-specific tensors (adjacency matrices, edge-type embeddings, sampled neighbor sets), and the locally held sub-graphs are highly non-IID. These characteristics can destabilize convergence and demand sophisticated aggregation strategies \cite{feddyn2023,DBLP:conf/icml/KarimireddyKMRS20}. On the other hand, tree-based ensembles, such as Gradient-Boosted Trees (GBTs), show comparable forecasting accuracy to GNNs \cite{DBLP:conf/nips/GrinsztajnOV22}, while having a simpler design and straightforward deployment in federated contexts \cite{DBLP:conf/eurosys/MaQBL23}.

    Building on this insight, we propose Bikelution, a framework that extends a proven GBT-based baseline \cite{DBLP:conf/edbt/TziorvasTT25} to the FL paradigm. In summary, the main contributions of this work are as follows:
    \begin{itemize}
        \item We propose an FL-enabled framework based on eXtreme Gradient Boosted Trees (XGBoost).
        \item We demonstrate the performance of Bikelution, in terms of prediction accuracy in mid-term prediction horizon up to six hours, using three real-world BSS datasets.
        \item We demonstrate the performance of Bikelution with respect to current-state-of-the-art.
        \item We study the effect of FL on the task at hand, i.e., the performance of Bikelution with respect to its centralized variant.
    \end{itemize} 
    Our experimental study across three real-world BSS datasets reveals that Bikelution can provide on-par prediction fidelity relative to its centralized variant, while preserving data-ownership requirements, thereby offering a scalable solution for next-generation bike-demand forecasting platforms. Compared to the current state-of-the-art \cite{DBLP:journals/tits/LiL24a}, Bikelution improves forecasting error up to 13.30\% and 14.96\%, in terms of MAE and RMSE, respectively.

    The rest of the paper is organized as follows. Section \ref{sec:RelatedWork} discusses related work. Section \ref{sec:ProblemFormulation} formulates the task at hand and presents the methodological architecture of our FL-enabled framework. Section \ref{sec:ExperimentalStudy} presents the core results of our experimental study. Finally, Section \ref{sec:Conclusions} concludes the paper, also giving hints for future work.

%% file: 2.RelatedWork.tex
 \section{Related Work}\label{sec:RelatedWork}
    Recent advances in bike-sharing demand forecasting reveal persistent challenges in modeling spatial-temporal dynamics and integrating heterogeneous external factors \cite{DBLP:journals/tsusc/PavlidisPYWMEKMD25}. Traditional approaches often assume static station networks, neglecting the time-varying connectivity and spatial heterogeneity inherent in bike-sharing systems (BSSs) \cite{ZhangSTResNet, LDBSS2020}. While external factors (e.g., points of interest, residential density) influence demand, many models treat these as auxiliary inputs rather than primary features, leading to suboptimal generalization. Furthermore, existing deep learning frameworks frequently fail to jointly capture local and long-range spatial dependencies, yielding fragmented predictions with limited operational utility.
    
    To address these limitations, recent work has progressively adopted dynamic graph topologies. Feng et al. \cite{FengSTAGNN2024} coupled a Dynamic Graph Attention Network (DGAT) with a Temporal Convolutional Network (TCN) to model fine-grained local dependencies, while a shared GNN-MLP module aggregated global spatial features.
    Similarly, Chen et al. \cite{ChenASTGCN2021} embedded a Temporal-Attention Module (TAM) within Spatial-Temporal Graph Convolutional Networks (ST-GCNs), generating probabilistic attention weights to emphasize salient temporal patterns before spatial convolution.
    Guo et. al.,\cite{GuoASTGCN2019} introduced ASTGCN, a novel graph convolution network based model that dynamically learns both spatial and temporal relationships across recent, daily, and weekly patterns, thereby overcoming these limitations. 
    In a similar line of research, Guo et al. \cite{GuoGGCN2024} proposed an edge-weighted Gated Graph Convolutional Network (GGCN), which derived edge weights from normalized station distance, travel duration, and demand-correlation scores to capture heterogeneous spatial relationships. Shao et al. \cite{ShaoSTHGCN2023} proposed STHGCN, a Spatial-Temporal Hierarchical Graph Convolutional Network, which fuses dual spatial embeddings via a gating mechanism to model temporal dependencies.
    
    Despite these advances, centralized data aggregation remains problematic due to privacy risks and difficulties in harmonizing heterogeneous municipal datasets. On the other hand, FL-based solutions focus on training centralized models using decentralized data \cite{DBLP:conf/aistats/McMahanMRHA17}, thereby, preserving data privacy and sovereignty. Cai and Liu \cite{CaiHSTFL2024} proposed a Heterogeneous Spatio-Temporal FL (HSTFL) architecture using vertical FL to resolve cross-domain feature heterogeneity, with a Cross-Client Virtual Node Alignment (VNA) module enabling privacy-preserving knowledge fusion. 
        
    Goto et al. \cite{DBLP:conf/smartcomp/GotoMRYY23} employ FedAvg for collaboratively training a fully-connected neural network across multiple service providers for taxi-demand prediction. Hu et al. \cite{DBLP:journals/tvt/HuYHLCYSAPW23} propose FedTDP for predicting short-term ride-sourcing demand using a spatio-temporal model (FedSTTDP) that combines Long Short-Term Memory (LSTM) to capture temporal correlations and Graph Convolutional Networks (GCN) for spatial correlations. 

    Closer to out work, Li and Liu \cite{DBLP:journals/tits/LiL24a} introduced MAFL, a novel Horizontal FL (HFL) framework for multimodal transport demand forecasting. Their approach utilizes a Fine-grained Graph Convolution Recurrent Network (F-GCRN) to capture dynamic spatiotemporal correlations and employs an Attentive Federated Learning mechanism to selectively merge model parameters based on layer-wise similarity. This work is considered, to the best of our knowledge, the state of the art in (short-term) BDF, with absolute error of 2.591 bike trips in 1 hr. prediction horizon. 
        
    In contrast to the above related work, Bikelution predicts future bike-sharing demand up to six hours, in one-hour intervals. By training gradient-boosted trees on locally partitioned data and aggregating only model updates, our approach attains MAE and RMSE comparable to its centralized baseline while preserving strict data-sovereignty guarantees. With respect to current state of the art, our solution improves forecasting fidelity up to 15\%, as demonstrated in Section \ref{subsec:ExperimentalResults}.

%% file: 3.ProposedApproach.tex
\section{Problem Formulation and Proposed Methodology}\label{sec:ProblemFormulation}
    In this section, we present the proposed Bikelution solution and describe it under both CML and FL paradigms.
    
    \begin{figure*}[!ht]
        \centering
        \includegraphics[width=\textwidth]{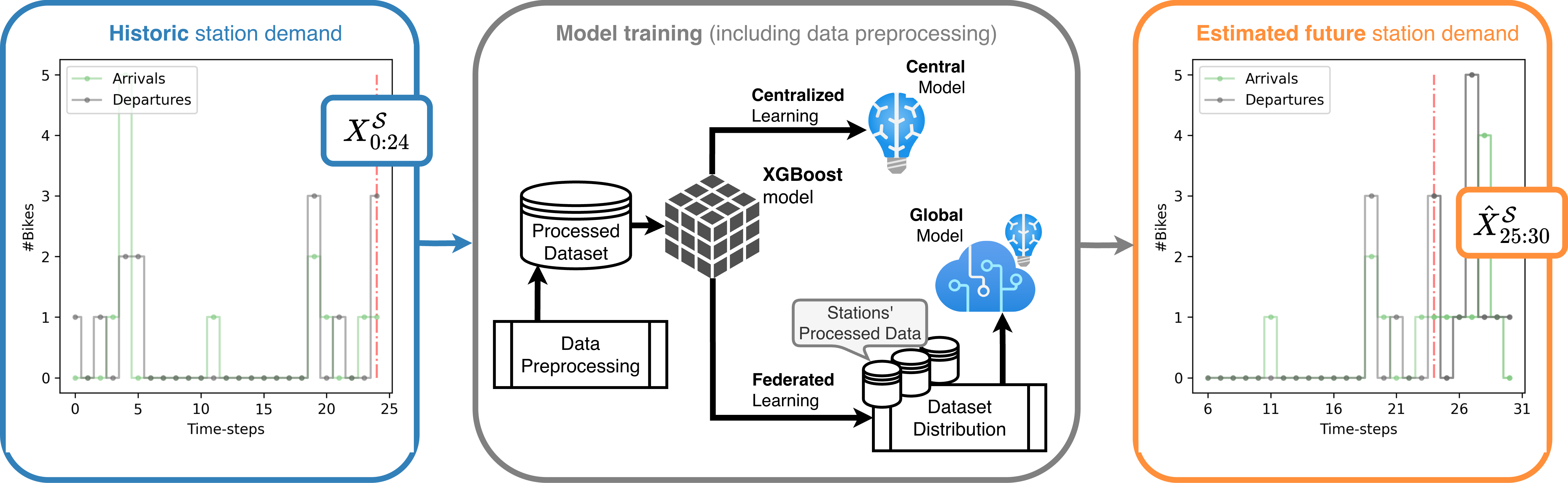}
        \caption{Overview of our Bikelution forecasting framework.}
        \label{fig:bikelution-overview}
    \end{figure*}
    
    \subsection{Problem Definition}\label{subsec:ProblemDefinition}
        Before we proceed to the actual formulation of the problem, we provide some preliminary definitions.

        \begin{definition}[Bike trip]\label{def:BicycleTrip}
            A bike trip $\mathcal{R}$ is defined as:
            \[
                \mathcal{R} = \langle \texttt{loc}_{start}, t_{start}, \texttt{loc}_{end}, t_{end} \rangle,
            \]
            where $\texttt{loc}_{start}$ and $\texttt{loc}_{end}$ denote the departing and arrival bike stations, respectively, and $t_{start}$ and $t_{end}$ are their corresponding timestamps.
        \end{definition}
        
        \begin{definition}[Bike demand]\label{def:BikeDemand}
            Given a temporal axis $\mathcal{T}$ with fixed sampling rate $\textit{sr}_{\text{freq}}$, let $\mathcal{T} = \{t_1, t_2, \ldots, t_n\}$ denote the discrete time intervals. For a bike station $\mathcal{S}$, the demand $D_t^\mathcal{S}$ during interval $t = \left[t_i, t_{i+1}\right)$ is defined as the 2D vector
            \[
                D_t^\mathcal{S} = 
                \begin{pmatrix}
                    A_t^\mathcal{S} \\
                    D_t^\mathcal{S}
                \end{pmatrix}
                =
                \begin{pmatrix}
                    \displaystyle\sum_{\mathcal{R}_k \in \mathcal{R}_{\mathcal{S}}} \mathbf{1}_{\{ \texttt{loc}_{end}^k = \mathcal{S} \ \land \ t_{end}^k \in \left[t_i, t_{i+1}\right) \}} \\[1.2ex]
                    \displaystyle\sum_{\mathcal{R}_k \in \mathcal{R}_{\mathcal{S}}} \mathbf{1}_{\{ \texttt{loc}_{start}^k = \mathcal{S} \ \land \ t_{start}^k \in \left[t_i, t_{i+1}\right) \}}
                \end{pmatrix},
            \]
            where $\mathcal{R}_{\mathcal{S}}$ denotes the set of all bike trips associated with station $\mathcal{S}$, and $\mathbf{1}_{\{\cdot\}}$ is the indicator function. The first component represents the number of arriving bikes (arrivals), and the second component represents the number of departing bikes (departures) during interval $t$.
        \end{definition}

        \begin{definition}[Bike demand forecasting]\label{def:BikeDemandForecasting}
            Given a dataset $X$ of historical bike demand time series for station $\mathcal{S}$, let 
            \[
                X_{t-\tau:t}^{\mathcal{S}} = \big\{ D_{t-\tau}^{\mathcal{S}}, D_{t-\tau+1}^{\mathcal{S}}, \ldots, D_{t-1}^{\mathcal{S}}, D_{t}^{\mathcal{S}} \big\}
            \]
            denote the input sequence over a look-back interval of length $\tau$ up to time $t$. The goal of bike demand forecasting is to train a data-driven model $F(\cdot)$ such that
            \[
                \hat{X}_{t+1:t+H}^{\mathcal{S}} = F\big( X_{t-\tau:t}^{\mathcal{S}} \big),
            \]
            where $\hat{X}_{t+1:t+H}^{\mathcal{S}} = \{ \hat{D}_{t+1}^{\mathcal{S}}, \hat{D}_{t+2}^{\mathcal{S}}, \ldots, \hat{D}_{t+H}^{\mathcal{S}} \}$ represents the estimated future demand over the forecast horizon $H$.
        \end{definition}

        If we recall Figure \ref{fig:bikelution-overview}, it provides an illustration of Definition \ref{def:BikeDemandForecasting}. At first, we aggregate the raw trip records of a bike station (cf., Definition \ref{def:BicycleTrip}), into a time series of arriving / departing bikes per fixed-size temporal interval, thereby obtaining the bike demand sequence described in Definition \ref{def:BikeDemand}.
        Afterwards, these sequences are enriched with additional features and fed into our Bikelution model, which is trained under both CML and FL paradigms. Finally, in the inference step, once we know the bike demand sequence $X_{\tau:t}^{\mathcal{S}}$ of station $\mathcal{S}$ up to interval $t$, we estimate its expected bike demand in the next $H$ intervals, or $X_{t+1:t+H}$. In our work, we set $\tau = 7$ days, $sr_{freq} = 1$ hour, and predict demand up to $H = 6$ intervals ahead. The details of our methodological framework and the respective workflow are provided in the sections that follow.

    \subsection{Extending to Federated Learning}\label{subsec:MethodologyFL}       
        Traditional centralized approaches often face challenges, such as privacy constraints and communication overhead. These issues are particularly significant in real-world BSSs, where historical bike demand data is inherently localized to individual entities and subject to privacy regulations. To address these challenges, we experiment with the FL paradigm, which enables collaborative model training across multiple participants without requiring raw data to be shared or centralized. This approach preserves privacy, minimizes communication costs, and potentially improves forecasting accuracy by leveraging the collective insights of distributed clients.

\begin{figure}[!ht]
            \centering
            \includegraphics[width=\columnwidth]{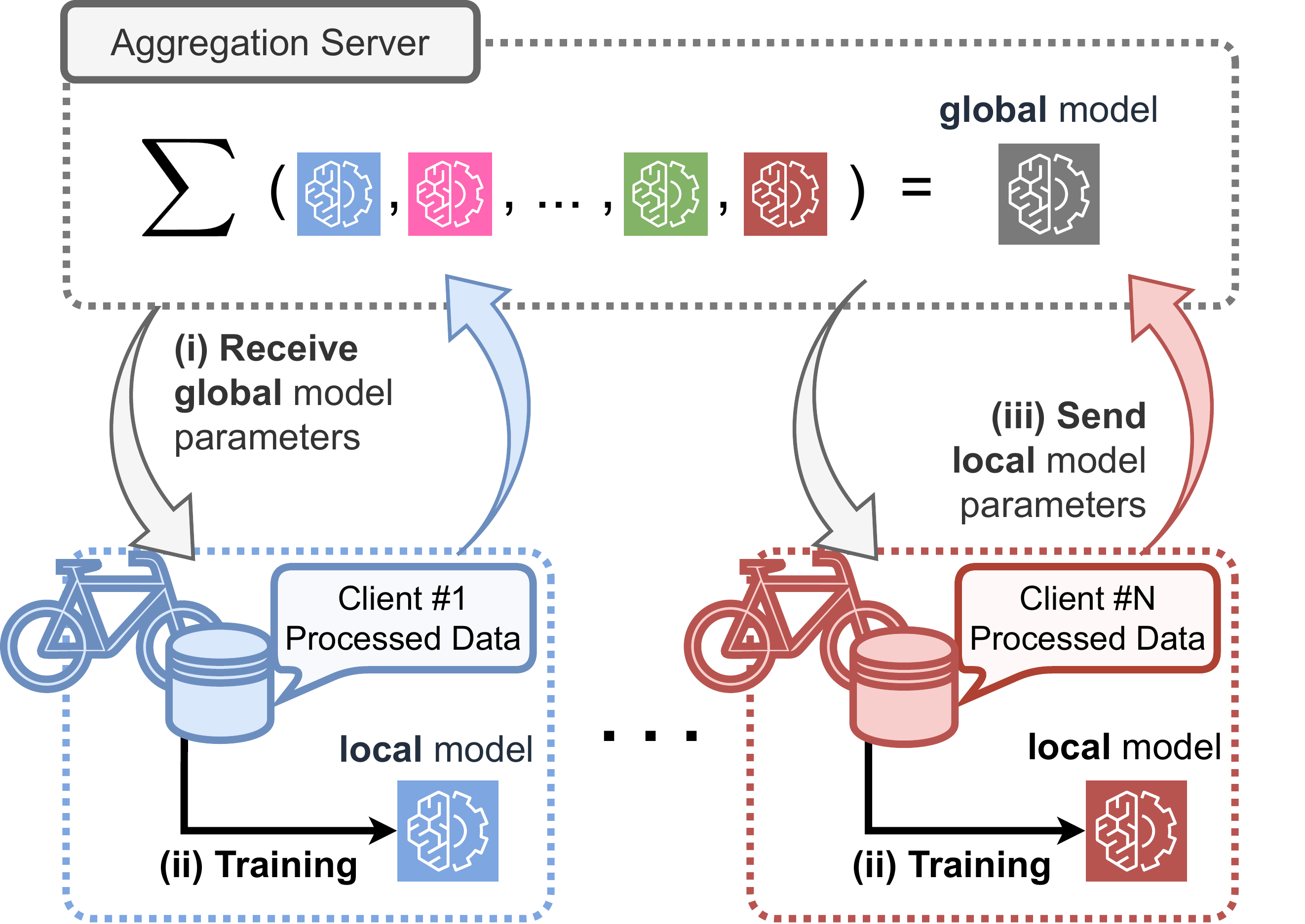}
            \caption{Bikelution Federated Workflow Overview}            
            \label{fig:federated-workflow}
        \end{figure}

        Figure \ref{fig:federated-workflow} illustrates our FL-based workflow. We define $N$ clients (i.e., BSSs), each containing historic bike demand data. To better understand how FL works, a round corresponds to a full cycle of global model distribution to clients and subsequent aggregation of updated local models, while an epoch denotes the number of iterations over the clients' local data during model training. At the start of the FL process, the aggregation server randomly selects one participant, and uses its local model as the initial global model. Afterwards, each subsequent FL round involves, (i) sending the parameters of the current global model to randomly selected clients in the federation with probability $p_{train}$, (ii) training the model locally on clients' corresponding data for a set number of local epochs, and (iii) receiving the newly trained local models in order to generate the updated global model.
        
        In our framework, each client is selected with probability $p_{train}=0.25$ and performs $E_{local} = 10$ local epochs per round with early stopping patience of $5$ epochs to mitigate overfitting \cite{DBLP:series/lncs/Prechelt12}, over $E_{global} = 15$ total rounds. For training the (local) Bikelution models, we adopt the same methodology as the centralized baselines. The (global)Bikelution model is optimized using FedProx \cite{DBLP:conf/mlsys/LiSZSTS20} with $\mu_{prox} = 0.125$. These hyperparameters ($p_{train}$, $E_{local}$, $E_{global}$ and $\mu_{prox}$) were empirically selected based on the values listed in \cite{DBLP:conf/mlsys/LiSZSTS20} and validated through convergence analysis of the (local)FedEDF models. 
        
        For the FL implementation of XGBoost, we adopt FedXGBllr \cite{DBLP:conf/eurosys/MaQBL23}, a novel horizontal FL framework that treats the learning rates of the aggregated tree ensembles as learnable parameters. This is implemented through a small one-layer 1D Convolutional Neural Network (1D-CNN) -based model, whose inputs correspond to the prediction outcomes of all trees from the aggregated ensemble. The kernel and stride size of the 1D-CNN are equal to the number of trees, $M$, per client, allowing each channel to represent learnable learning rates.

        The process involves clients initially training local XGBoost tree ensembles, which are then sorted, aggregated and broadcast by the server, followed by clients collaboratively training the 1D-CNN model with the prediction outcomes of all trees from these aggregated ensembles on their local data samples serving as inputs. 
        To keep communication overhead manageable and to facilitate comparison with the centralized variant, we set $M=37$ estimators per client. 

%% file: 4.ExperimentalSetup.tex
\section{Experimental Study}\label{sec:ExperimentalStudy}
    In this section, we compare the efficiency of Bikelution using three real-world shared mobility datasets, and present our experimental results. The models were implemented in Python using XGBoost\footnote{XGBoost: An optimized distributed gradient boosting library. \url{https://xgboost.readthedocs.io/en/stable/}} and PyTorch\footnote{PyTorch: An Imperative Style, High-Performance Deep Learning Library. \url{https://pytorch.org}}, and trained in a federated manner via Flower\footnote{Flower: A Friendly Federated Learning Framework. \url{https://flower.dev}} on a 256-CPU, 1 TB RAM workstation. In addition, we provide our study as open-source\footnote{The corresponding source code used in our experimental study is available at: \url{https://github.com/DataStories-UniPi/Bikelution}} in order for the researchers and practitioners in the field to take the most benefit from it.

    \subsection{Experimental Setup, Datasets and Preprocessing}\label{subsec:ExperimentalSetup}
        For the purposes of our experimental study, we use three open real-world shared mobility demand datasets, hereafter referred to as ``NYC''\footnote{The dataset is publicly available at: \url{https://citibikenyc.com/system-data}}, ``Chicago''\footnote{The dataset is publicly available at: \url{https://divvybikes.com/system-data}}, and ``Barcelona''\footnote{The dataset is publicly available at: \url{https://doi.org/10.5281/zenodo.17650616}.}. 
        
        All raw records undergo a uniform cleaning routine: we drop records with missing fields, trip duration greater than 24 hours or unrealistically short round-trips, as well as stations with less than 3 rentals per day \cite{FengSTAGNN2024,DBLP:conf/edbt/TziorvasTT25}. Afterwards, we create the bike-demand values defined in Definition \ref{def:BikeDemand} by aggregating raw trips per station into a two-dimensional time-series sampled at one-hour intervals, which records the numbers of arriving and departing bikes.

        \begin{figure*}[!ht]
            \centering
            \includegraphics[width=\textwidth]{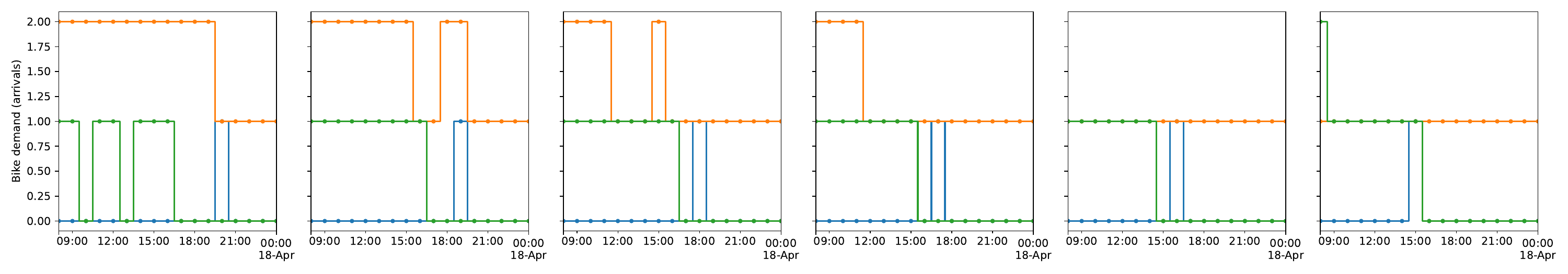}\\
            \includegraphics[width=\textwidth]{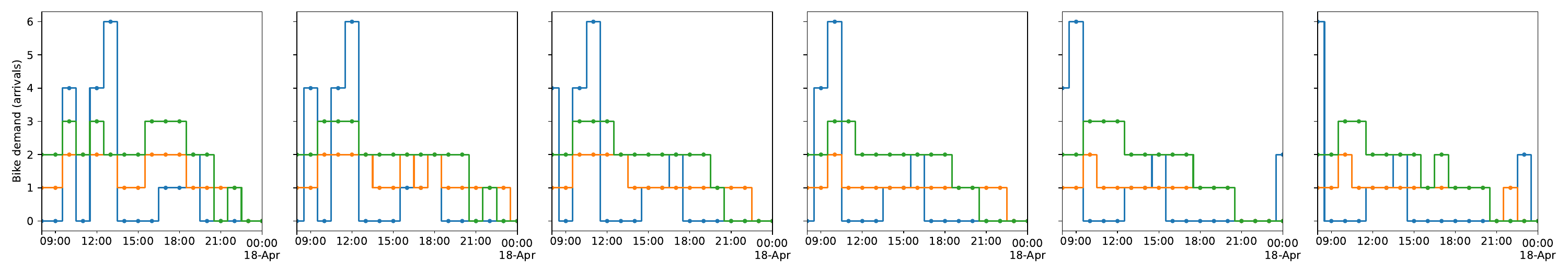}\\
            \includegraphics[width=\textwidth]{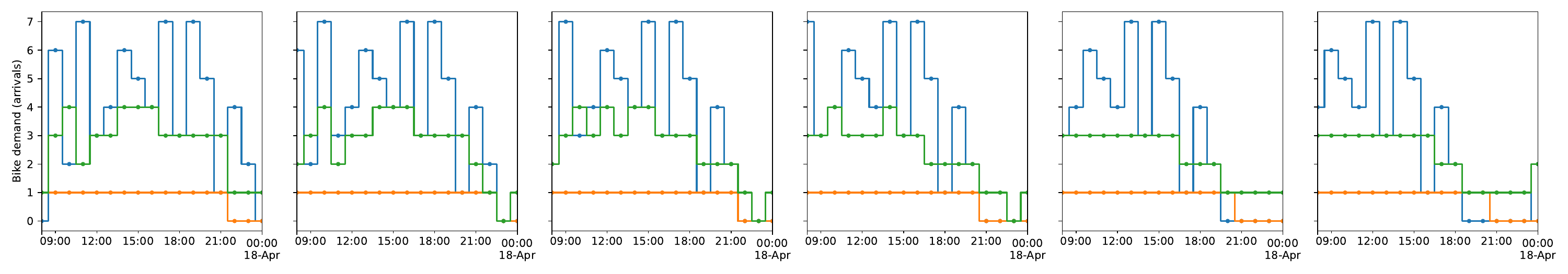}
            \caption{Bike demand forecasting (arrivals) at three representative stations on the test set of the NYC dataset during March 17, 2022. Each row corresponds to bike stations whose HFL-global RMSE falls at the 25\textsuperscript{th} (top), 50\textsuperscript{th} (middle), and 75\textsuperscript{th} (bottom) percentiles. Each column corresponds to one of the six prediction horizons (left-to-right: 1-6 hours). Solid blue lines represent the actual demand, while the green and orange lines show the predictions of the CML and HFL-global models, respectively.}
            \label{fig:nyc-bike-arrivals}
        \end{figure*}
        
        Beyond raw temporal and contextual signals, we engineer a rich feature representation that captures multi-scale temporal dependencies, trend dynamics, and seasonal patterns inherent in spatial demand data. For each spatial entity (i.e., bike station), we construct a compact yet informative feature set organized into four complementary groups.

        We incorporate fundamental temporal identifiers that encode the cyclical structure of human activity patterns: hour-of-day $(h \in {0,1,\ldots,23})$, day-of-week $(d \in {0,1,\ldots,6})$, and weekly seasonality indicators. Rather than using simple one-hot or sinusoidal encoding, we apply a Radial Basis Function (RBF) kernel encoding scheme to map each temporal index onto a continuous feature space. 
        Specifically for the hour-of-day we compute:
        \[
            \phi_h^{(j)} = \exp\left(-\frac{(h - \mu_h^{(j)})^2}{2\sigma_h^2}\right), \quad j = 1, 2, \dots, K_h
        \]
        where $K_h$ denotes the number of RBF centers, $\mu_h^{(j)}$ represents the center of the j-th basis function (evenly spaced across the 24-hour cycle) and $\sigma_h$ controls the bandwidth.
        This encoding preserves temporal periodicity, while enabling the model to learn non-linear periodic patterns through a more flexible representation than rigid sinusoidal projections. 

        Auto-regressive dynamics and short-term dependencies, are captured through lagged demand values at multiple horizons
        To account for exogenous factors shaping demand, we include binary indicators for public holidays (national and regional), typical work hours (e.g., 09:00–17:00 on weekdays), and school schedules where applicable. These covariates are derived from authoritative calendar sources and encoded as binary vectors aligned with the temporal resolution of the target series. Rolling statistics (e.g., rolling mean, and minima/maxima) provide aggregated views of recent demand trends and capture local volatility and range-bound behavior. Complementing those features, we employ exponentially weighted moving (EWM) features, which assign decaying weights to past observations. This formulation offers a smoother, more responsive alternative to fixed-window rolling statistics, particularly beneficial in non-stationary demand conditions.

        To explicitly model multi-periodic patterns beyond the weekly cycle, we apply a Fourier-based seasonal decomposition using a subset period of historical observations as a burn-in set. Specifically, for each entity we compute the Fourier components:
        \[
            F_k(t) = a_k \cos\left(\frac{2\pi k t}{T}\right) + b_k \sin\left(\frac{2\pi k t}{T}\right),
        \]
        where $T$ denotes the period, and $k$ indexes the harmonic order. Coefficients $(a_k, b_k)$ are estimated from the burn-in set comprising the first $N$ observations. The reconstructed seasonal signal is then included as a deterministic feature, providing the model with an explicit representation of periodic structure that is robust to missing data in the recent window.
        
        \input{tabs/tab-dataset-description}

        \input{tabs/tab-bikelution-results}

        To assess the quality of our models we employ four well-known metrics, namely, Symmetric Mean Absolute Percentage Error (SMAPE), Mean Arctangent Absolute Percentage Error (MAAPE), Mean Absolute Error (MAE) and Root Mean Squared Error (RMSE). Each metric highlights a different aspect of forecasting performance for the next 6-hour prediction window.

        In particular, SMAPE and MAAPE translate errors into percentage terms that are readily interpretable, while MAE and RMSE quantify absolute deviations in the unit of measure of the forecast (i.e., number of arriving / departed bikes). The inherent weighting toward specific error characteristics (e.g., outliers, low-demand periods, etc.) of each metric ensures that our assessment is robust across the diverse operating conditions typical of BSS demand.

    \subsection{Experimental Results}\label{subsec:ExperimentalResults}
        After the preprocessing steps described in Section \ref{subsec:ExperimentalSetup}, we split the temporal axis of our data into training, validation, and test sets using a ratio of 70:20:10\% (1 fold). For each data source, we train two Bikelution instances, one in a fully centralized setting as described in \cite{DBLP:conf/edbt/TziorvasTT25}, and another in a horizonal FL setting as described in Section \ref{subsec:MethodologyFL}. Table \ref{tab:bikelution-results} illustrates the performance of each variant across six one-hour prediction horizon look-aheads.

        The evaluation spans six prediction horizons (1–6 steps ahead) and employs four standard error metrics, namely SMAPE (\%), MAAPE (radians), MAE (bikes), and RMSE (bikes). Values are reported as arrivals / departures. Centralized (CML) consistently outperforms HFL across all datasets and metrics. The performance gap is most pronounced for the Chicago dataset, where HFL-global SMAPE values exceed CML by approximately 25–35\% across horizons. Error increases with prediction horizon for both paradigms, reflecting the inherent uncertainty in multi-step forecasting. However, in some cases CML exhibits a steeper error gradient than HFL-global, suggesting that the FL model may generalize more stably across horizons despite higher absolute error.

        Focusing on the NYC dataset, Figure \ref{fig:nyc-bike-arrivals} showcases the predictions of Bikelution in both CML and HFL-global configurations across three representative stations. These stations were selected to represent the 25\textsuperscript{th}, 50\textsuperscript{th}, and 75\textsuperscript{th} percentiles of forecasting error (RMSE), respectively. As illustrated, our model accurately captures average bike demand, with the CML model prediction fidelity being slightly more accurately than HFL-global, as observed in Table \ref{tab:bikelution-results}. However, for stations with higher demand, the model defaults to a constant minimal value. This behavior is attributed to client drift \cite{DBLP:conf/icml/KarimireddyKMRS20}, which can be mitigated by fine-tuning the aggregation strategy on the target dataset or employing more advanced federated optimization strategies \cite{feddyn2023,DBLP:conf/icml/KarimireddyKMRS20}.
        
        These results demonstrate that horizontal federated learning achieves competitive forecasting accuracy compared to centralized training while preserving data sovereignty by keeping raw observations localized to respective clients. This trade-off is favorable in privacy-sensitive bike-sharing deployments where cross-jurisdictional data sharing is constrained by regulatory or operational barriers.

        Since the source code for MAFL \cite{DBLP:journals/tits/LiL24a} is not publicly available, we restrict our reproducibility study to the the first (1-step-ahead) forecasting horizon of the NYC dataset. Compared with F-GCRN, the CML-based variant of Bikelution reduces MAE by 20.68\% and RMSE by 25.11\%. Moreover, (global)Bikelution outperforms MAFL, achieving a 13.30\% reduction in MAE and a 14.96\% reduction in RMSE. This is a key result of our study, that further confirms the value of FL and strengthens the position of our work in privacy-aware bike-demand forecasting. 
        

%% file: tabs/tab-dataset-description.tex
\begin{table}[!ht]
    \renewcommand{\arraystretch}{1.8}
    \centering
    \caption{Statistics of the datasets used in our experimental study, after the preprocessing phase.}
    \label{tab:dataset-description}
    \resizebox{\columnwidth}{!}{%
        \begin{tabular}{@{}lcccc@{}}
            \toprule
            Dataset         
            & Time Span     
            & \#Trips 
            & \#Samples 
            & \#Stations 
            \\
            \midrule
            Barcelona
            & \makecell[c]{2020-03-30 --\\2024-12-30}
            & 80,118,067
            & 125,757,576
            & 504
            \\
            NYC
            & \makecell[c]{2021-02-16 --\\2022-05-31}
            & 36,323,801
            & 101,195,703
            & 1497
            \\
            Chicago
            & \makecell[c]{2021-02-16 --\\2022-05-31}
            & 7,104,591
            & 28,594,377
            & 423
            \\ 
            \bottomrule
        \end{tabular}%
    }
\end{table}

%% file: tabs/tab-bikelution-results.tex
\begin{table*}[ht]
    \centering
    \caption{Accuracy of Bikelution forecasts (arrivals / departures) per dataset (lower is better).}
    \label{tab:bikelution-results}
    \renewcommand{\arraystretch}{1.3}
    \begin{tabular}{ccrrrrrrr}
        \toprule
        &
        & Metric 
        & Horizon 1 
        & Horizon 2 
        & Horizon 3 
        & Horizon 4 
        & Horizon 5 
        & Horizon 6 \\
        \midrule
        \multirow{9}{*}{\rotatebox{90}{Barcelona}}
        % ---------- centralized ----------
        & \multirow{4}{*}{\rotatebox{90}{CML}}
        & SMAPE (\%)
        & 0.691 / 0.693 
        & 0.743 / 0.754
        & 0.767 / 0.779
        & 0.777 / 0.789 
        & 0.785 / 0.792 
        & 0.791 / 0.799
        \\
        &
        & MAAPE (rads)
        & 0.563 / 0.570 
        & 0.612 / 0.623
        & 0.629 / 0.641 
        & 0.637 / 0.648
        & 0.641 / 0.647 
        & 0.642 / 0.649
        \\
        &
        & MAE (bikes) 
        & 1.930 / 1.921
        & 2.111 / 2.117 
        & 2.183 / 2.194 
        & 2.213 / 2.224 
        & 2.238 / 2.249 
        & 2.269 / 2.283
        \\
        &
        & RMSE (bikes) 
        & 2.741 / 2.712 
        & 2.955 / 2.946
        & 3.030 / 3.031 
        & 3.060 / 3.066 
        & 3.092 / 3.100 
        & 3.134 / 3.142
        \\
        \addlinespace
        % ---------- global_fedxgb ----------
        & \multirow{4}{*}{\rotatebox{90}{HFL-global}}
        & SMAPE (\%) 
        & 0.843 / 0.852 
        & 0.873 / 0.880 
        & 0.887 / 0.894 
        & 0.893 / 0.899 
        & 0.893 / 0.899 
        & 0.895 / 0.901
        \\
        &
        & MAAPE (rads)
        & 0.715 / 0.731
        & 0.727 / 0.742
        & 0.730 / 0.745
        & 0.730 / 0.745 
        & 0.730 / 0.743 
        & 0.732 / 0.744
        \\
        &
        & MAE (bikes) 
        & 2.092 / 2.133 
        & 2.336 / 2.361
        & 2.439 / 2.475 
        & 2.480 / 2.512 
        & 2.495 / 2.517 
        & 2.518 / 2.539 
        \\
        &
        & RMSE (bikes) 
        & 2.881 / 2.868 
        & 3.257 / 3.207 
        & 3.400 / 3.361 
        & 3.449 / 3.414 
        & 3.469 / 3.435 
        & 3.499 / 3.470
        \\
        \midrule
        \multirow{9}{*}{\rotatebox{90}{NYC}}
        % ---------- centralized ----------
        & \multirow{4}{*}{\rotatebox{90}{CML}}
        &  SMAPE (\%)
        & 0.997 / 1.036 
        & 1.018 / 1.061 
        & 1.040 / 1.087 
        & 1.048 / 1.095 
        & 1.048 / 1.092 
        & 1.047 / 1.091
        \\
        &
        & MAAPE (rads) 
        & 0.761 / 0.800
        & 0.778 / 0.819
        & 0.796 / 0.842
        & 0.803 / 0.847 
        & 0.804 / 0.845 
        & 0.804 / 0.844 
        \\
        &
        & MAE (bikes) 
        & 2.067 / 2.099 
        & 2.097 / 2.134
        & 2.124 / 2.161
        & 2.135 / 2.172
        & 2.146 / 2.178
        & 2.159 / 2.192
        \\
        &
        & RMSE (bikes) 
        & 2.948 / 2.949
        & 2.974 / 2.978
        & 2.995 / 2.994 
        & 3.003 / 2.999 
        & 3.012 / 3.007 
        & 3.027 / 3.020
        \\
        \addlinespace
        % ---------- global_fedxgb ----------
        & \multirow{4}{*}{\rotatebox{90}{HFL-global}}        
        & SMAPE (\%)      
        & 1.097 / 1.107
        & 1.128 / 1.142 
        & 1.178 / 1.191 
        & 1.207 / 1.212 
        & 1.214 / 1.205
        & 1.237 / 1.224
        \\
        &
        & MAAPE (rads) 
        & 0.718 / 0.720
        & 0.735 / 0.740
        & 0.765 / 0.771 
        & 0.776 / 0.774 
        & 0.772 / 0.758 
        & 0.789 / 0.771 
        \\
        &
        & MAE (bikes) 
        & 2.250 / 2.243
        & 2.255 / 2.252
        & 2.261 / 2.257 
        & 2.263 / 2.257 
        & 2.264 / 2.255 
        & 2.276 / 2.266
        \\
        &
        & RMSE (bikes) 
        & 3.319 / 3.273
        & 3.325 / 3.283
        & 3.333 / 3.287
        & 3.341 / 3.298
        & 3.351 / 3.307
        & 3.359 / 3.314 
        \\
        \midrule
        \multirow{9}{*}{\rotatebox{90}{Chicago}}
        % ---------- centralized ----------
        & \multirow{4}{*}{\rotatebox{90}{CML}}
        & SMAPE (\%)
        & 1.016 / 0.992
        & 1.040 / 1.025
        & 1.079 / 1.053
        & 1.111 / 1.078
        & 1.124 / 1.081
        & 1.124 / 1.086
        \\
        &
        & MAAPE (rads)
        & 0.769 / 0.748
        & 0.785 / 0.771 
        & 0.814 / 0.792 
        & 0.840 / 0.812 
        & 0.849 / 0.813 
        & 0.846 / 0.815
        \\
        &
        & MAE (bikes) 
        & 1.265 / 1.246
        & 1.285 / 1.269
        & 1.305 / 1.286 
        & 1.320 / 1.298 
        & 1.326 / 1.303
        & 1.331 / 1.313
        \\
        &
        & RMSE (bikes) 
        & 1.956 / 1.938 
        & 1.976 / 1.958
        & 1.990 / 1.972
        & 1.996 / 1.981
        & 2.005 / 1.986 
        & 2.013 / 1.997
        \\
        \addlinespace
        % ---------- global_fedxgb ----------
        & \multirow{4}{*}{\rotatebox{90}{HFL-global}}
        & SMAPE (\%) 
        & 1.369 / 1.366
        & 1.369 / 1.366
        & 1.369 / 1.366
        & 1.370 / 1.368 
        & 1.370 / 1.368 
        & 1.371 / 1.369
        \\
        &
        & MAAPE (rads) 
        & 1.061 / 1.057 
        & 1.059 / 1.056
        & 1.052 / 1.049
        & 1.045 / 1.042
        & 1.043 / 1.040
        & 1.041 / 1.039
        \\
        &
        & MAE (bikes) 
        & 1.555 / 1.546
        & 1.556 / 1.551 
        & 1.517 / 1.513 
        & 1.480 / 1.475 
        & 1.467 / 1.464 
        & 1.462 / 1.460
        \\
        &
        & RMSE (bikes) 
        & 2.212 / 2.201
        & 2.216 / 2.208 
        & 2.180 / 2.173 
        & 2.151 / 2.144 
        & 2.142 / 2.137
        & 2.139 / 2.136
        \\
        \bottomrule
    \end{tabular}
\end{table*}

%% file: 5.Conclusions.tex
\section{Conclusions}\label{sec:Conclusions}
    In this paper, we presented Bikelution, a horizontal federated learning extension of the MoDE-Boost model \cite{DBLP:conf/edbt/TziorvasTT25} for bike demand forecasting, which improves forecasting error over the current state of the art up to 14.96\%, in terms of RMSE. 
    In particular, through an extensive experimental study on three real-world BSS datasets, we demonstrated that the federated approach achieves on-par predtiction fidelity with respect to the centralized baseline, with moderate performance degradation.
    
    In the near future, we aim to further adjust the architecture of the Bikelution model by incorporating additional external factors. In a parallel line of research, we aim to address the client-drift issue of Bikelution by incorporating advanced optimization strategies \cite{feddyn2023,DBLP:conf/icml/KarimireddyKMRS20} and assess their pros and cons with respect to the current approach.